\documentclass{article}

     \usepackage[preprint]{neurips_2019}

\usepackage[utf8]{inputenc} 
\usepackage[T1]{fontenc}    
\usepackage{hyperref}       
\usepackage{url}            
\usepackage{booktabs}       
\usepackage{amsfonts}       
\usepackage{amsmath}
\usepackage{nicefrac}       
\usepackage{microtype}      
\usepackage{algorithm}
\usepackage{algorithmic}
\usepackage{graphicx}
\usepackage{epstopdf}
\usepackage{subfigure}

\title{Three-Stage Subspace Clustering Framework with Graph-Based Transformation and Optimization}

\author{%
  Shuai Yang,\qquad Wenqi Zhu,\qquad Yuesheng Zhu\\
  Communication and Information Security Lab, \\Institute of Big Data Technologies,\\
  Shenzhen Graduate School, Peking University\\
  \texttt{ethanyang@pku.edu.cn \quad wenqizhu@pku.edu.cn \quad zhuys@pkusz.edu.cn} \\
}

\begin{document}

\maketitle

\begin{abstract}
  Subspace clustering (SC) refers to the problem of clustering high-dimensional data into a union of low-dimensional subspaces.  Based on spectral clustering, state-of-the-art approaches solve SC problem within a two-stage framework. In the first stage, data representation techniques are applied to draw an affinity matrix from the original data. In the second stage, spectral clustering is directly applied to the affinity matrix so that data can be grouped into different subspaces. However, the affinity matrix obtained in the first stage usually fails to reveal the authentic relationship between data points, which leads to inaccurate clustering results. In this paper, we propose a universal Three-Stage Subspace Clustering framework (3S-SC). Graph-Based Transformation and Optimization (GBTO) is added between data representation and spectral clustering. The affinity matrix is obtained in the first stage, then it goes through the second stage, where the proposed GBTO is applied to generate a reconstructed affinity matrix with more authentic similarity between data points. Spectral clustering is applied after GBTO, which is the third stage. We verify our 3S-SC framework with GBTO through theoretical analysis. Experiments on both synthetic data and the real-world data sets of handwritten digits and human faces demonstrate the universality of the proposed 3S-SC framework in improving the connectivity and accuracy of SC methods based on $\ell_0$, $\ell_1$, $\ell_2$ or nuclear norm regularization.
\end{abstract}

\section{Introduction}

High-dimensional data, such as images and documents, are ubiquitous in many applications of computer vision, e.g., face clustering [1], image representation and compression [2], and motion segmentation [3,4]. In order to deal with these data in a high-dimensional ambient space, a union of low-dimensional data is adopted to approximate the original high-dimensional data, which is known as Subspace Clustering (SC) [5]. The task of SC is to partition the data into groups so that data points in a group come from the same subspace.

Many methods for subspace clustering have been proposed, including algebraic [6], iterative [7]-[10], statistical [11]-[16], and spectral clustering based methods [17]-[27]. State-of-the-art methods tend to adopt spectral clustering due to its advantage to reveal the relationship between data points based on graph.
In spectral clustering, data points are viewed as vertexes in set $X = \left[ x_1, …,x_N \right]\in \mathbb{R}^{D\times N}$ in an undirected graph $G = \left( X, E \right)$, and the task is to partition $G$ into different subgraphs. Usually, an affinity matrix (or similarity matrix) $W$ which has the similarity between each vertex pair as its elements is generated to represent $G$. By applying spectral clustering to $W$, the data points will be clustered into different subspaces precisely.

Spectral clustering based methods usually follow a two-stage framework. In the first stage, with data representation techniques, a coefficient matrix $C$ is generated from the original data based on the self-expressiveness property of data belonging to the same subspace, and then an affinity matrix $W$ is obtained from $C$. That is,

\begin{equation}
x_j = Xc_j,\ c_{jj} = 0,\quad \mbox{or equivalently}\quad X = XC,\ diag(C) = 0,
\end{equation}
\begin{equation}
w_{ij} = \left|c_{ij}\right| + \left|c_{ji}\right| \quad \mbox{or equivalently}\quad  W = |C| + |C^T|,
\end{equation}

where $X = [x_1, …,x_N]\in \mathbb{R}^{D\times N}$is the data matrix, $C = [c_1, …,c_N]\in \mathbb{R}^{D\times N}$ is the coefficients matrix, and $W$ is the affinity matrix. Then spectral clustering is directly applied to $W$ to partition data points into subspaces in the second stage.

However, the affinity matrix $W$ obtained in the first stage of the two-stage framework contains insufficient information of $G$ since it's usually sparse or low rank, which is obviously not capable to represent the relationship between data points and inappropriate to be applied with spectral clustering directly. Applying spectral clustering to $G$ will lead to inaccurate results especially in some data sets with large number of data points. Two factors lead to this problem. Firstly, too much $zero$ elements in $W$ deny potential relations among data points, which makes the number of connections between data points particularly small. Secondly, nonzero elements in $W$ cannot reveal the authentic similarity between data points.

In this paper, we propose a universal Three-Stage Subspace Clustering framework (3S-SC), which aims to overcome the drawbacks of the two-stage SC framework. It provides a Graph-Based Transformation and Optimization (GBTO) mechanism which turns the original $W$ obtained from data representation techniques into an optimized affinity matrix $W^*$ which is more capable to represent the distribution of data points in the high-dimensional ambient space. In GBTO, we adopt the classic Floyd-Warshall algorithm as an optimization strategy to solve this SC problem. Besides, depending on different application scenarios, we propose two implementations of the 3S-SC framework which are Hard 3S-SC and Soft 3S-SC.

In addition, we compare and analyze different transformation strategies in GBTO which determines how weight and distance are transformed into each other. Finally, we set experiments on synthetic data and real-world data sets of handwritten digits as well as human faces with varying lighting to verify our theoretical analysis. The universality of the 3S-SC framework in improving the accuracy and connectivity of SC methods with different norm regularizations is demonstrated in comparison with traditional two-stage framework.

\paragraph{Related work.} Current spectral clustering based methods tend to apply $\ell_0$, $\ell_1$, $\ell_2$ or nuclear norm regularization on the coefficient matrix $C$, including original Sparse Subspace Clustering using $\ell_1$ norm (SSC-$\ell_1$) [28][29], Least Squares Regression (LSR) [17] using $\ell_2$ norm, Low Rank Representation (LRR) [30][31] using nuclear norm, Elastic Net Subspace Clustering (ENSC) [27] using a mixture of $\ell_1$ and $\ell_2$ norm and SSC by Orthogonal Matching Pursuit (OMP) [26] using $\ell_0$ norm. Meanwhile, Block Diagonal Representation (BDR) [32] uses block diagonal matrix induced regularizer to directly pursue the block diagonal matrix. Such methods divide the SC problem into two steps described as two-stage SC framework. Though Structured Sparse Subspace Clustering (SSSC) [33] uses a joint affinity learning and subspace clustering framework to re-weight the representation matrix, it still ignores that $W$ generated by (SSC-$\ell_1$) in the first stage cannot represent the authentic distribution of high-dimensional data points. Our paper presents a more universal framework which can be applied to SC methods with different data representation techniques in the first stage.

\section{Three-Stage Subspace Clustering: A universal framework}

The main difference between these spectral clustering based SC methods lies in how the affinity matrix is obtained. Therefore, these algorithms can be concluded in a Two-Stage Subspace Clustering framework. In order to deal with its drawbacks mentioned in Section 1.1, we propose a Three-Stage Subspace Clustering, which is universal for state-of-the-art SC algorithms. Step 3 in Algorithm 1 is added in comparison with two-stage SC. As is shown in Figure 1, 3S-SC contains Graph-Based Transformation and Optimization, details of which is in Section 3. Depending on the application scenario, two implementations of the 3S-SC framework are proposed:

\begin{itemize}
\item Hard 3S-SC: in this case all elements in affinity matrix $W$ are applied with Graph-Based transformation and optimization, in which way the number of nonzero entries in $W$ will increase. It works well especially in data sets with plenty subspaces and data points.

\item Soft 3S-SC: in this case only nonzero elements in W are applied with GBTO, which will not change the sparsity in $W$. It just reconstructs the existing relations by adjusting similarity.
\end{itemize}


\begin{algorithm}[h]
\caption{Three-Stage Subspace Clustering}
\label{alg:2}
\begin{algorithmic}[1]
\REQUIRE ~~\\ 
Dataset $X = [x_1, …,x_N]\in \mathbb{R}^{D\times N}$, data representation technique $T$
\STATE Generate coefficients matrix $C$ by using $T$ \qquad {\it$\backslash\backslash$start of the first stage}
\STATE Generate affinity matrix by $W = \left|C\right| + \left|C^T\right|$  or half\qquad {\it$\backslash\backslash$end of the first stage}
\STATE Generate $W^*$ by graph-based transformation and optimization (GBTO){\it$\quad\backslash\backslash$the second stage}
\STATE Apply spectral clustering to the affinity matrix $W$ {\it\qquad$\backslash\backslash$the third stage}
\ENSURE ~~\\ 
 Clustering results $S$
\end{algorithmic}
\end{algorithm}

\begin{figure}[b]
  \centering
  \includegraphics[width=13.9cm]{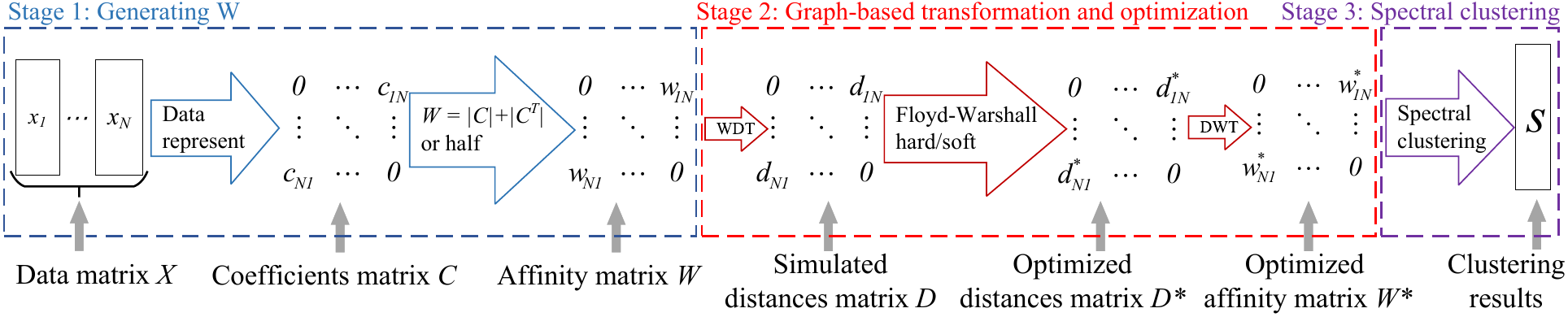}\\
  \caption{3S-SC. Graph-based transformation and optimization is applied in 3S-SC.}\label{fig}
\end{figure}

\section{Graph-based Transformation and Optimization (GBTO)}
In the last stage of 3S-SC, we apply spectral clustering to get SC results. To accomplish this, an affinity (or similarity) matrix $W\in\mathbb{R}^{N\times N}$ containing sufficient information of how high-dimensional data points distribute needs to be generated. As described in Section 1, the affinity matrix $W\in \mathbb{R}^{N\times N}$ obtained in the first stage is usually sparse or low rank, which is obviously unqualified to represent the relationship between data points and inappropriate to be applied with spectral clustering directly. To solve this problem, Graph-based Transformation and Optimization algorithm (GBTO) is proposed.

\begin{algorithm}
\caption{Graph-based Transformation and Optimization}
\label{alg:3}
\begin{algorithmic}[1]
\REQUIRE ~~\\ 
Original affinity matrix $W\in \mathbb{R}^{N\times N}$ learned in the first stage of 3S-SC
\STATE Transform $W$ into simulated distances matrix $D\in \mathbb{R}^{N\times N}$
\STATE Generate $D^*\in \mathbb{R}^{N\times N}$ by minimizing elements in $D$
\STATE Transform $D$ into a new affinity matrix $W^*\in \mathbb{R}^{N\times N}$
\ENSURE ~~\\ 
$W^*\in \mathbb{R}^{N\times N}$ which shows a more authentic relationship between data points
\end{algorithmic}
\end{algorithm}

\paragraph{Definition 1 (Simulated distances and simulated graph)} \emph{Elements in matrix $D$ generated from the affinity matrix $W$ to represent distances between data points are called simulated distances and $D$ is called simulated distances matrix, while the graph containing data points with simulated distances is defined as simulated graph.}

Stage 2 in Figure 1 describes how GBTO works. By transforming the affinity matrix $W$ into simulated distances matrix $D\in \mathbb{R}^{N\times N}$ via Weight-Distance Transformation (WDT), a simulated undirected graph $G$ is generated. Then $D^*\in \mathbb{R}^{N\times N}$ with minimized distances is generated from $D$ by applying Floyd-Warshall, after which a new affinity matrix $W^*$ with optimized similarity is generated by applying Distance-Weight Transformation (DWT) to $D^*$.

Obviously, GBTO can be divided into two parts: transformation between weights and distances (section 3.1), and optimization for simulated distances which determines how the relationship between data points is reconstructed (section 3.2).

\subsection{Transformation strategy between weights and distances (WDT and DWT)}
In an undirected graph $G = (X, E)$ with vertex set $X$, the weight (or similarity) between two data points grows approximately in inverse proportion to the distance between them, and the transformation should obey this rule as well. Due to the column normalization, the weight are distributed in the range $[0,1]$. Thus WDT and DWT can be approximated as $f(x)=\tfrac{1}{x} (0<x<1)$, where $x$ here stands for weights. Both f(x) and its inverse transformation $f^{-1}(x)$ are simple in computation. As is shown in Figure 2, compared with another proposed transformation $g(x) =1-$ln$x (0<x<1)$, $f(x)$ enjoys more stable changes when $x$ approaches $zero$, which means the $f(x)$ transformation pays more attention to small weights. It's a better mapping for GBTO aiming to deal with abnormally small similarity which is irrational in ternary relationship as described later in (6) and (7). Thus, transformation algorithms between affinity matrix and distances matrix are proposed as below.

\begin{algorithm}
\caption{Weight-Distance Transformation (WDT)}
\label{alg:4}
\begin{algorithmic}[1]
\REQUIRE ~~\\ 
An affinity matrix $W\in \mathbb{R}^{N\times N}$
\STATE $d_{ij} = \tfrac{1}{w_{ij}}$ if $w_{ij} \not= 0$\quad or\quad $d_{ij} = \infty$ if $w_{ij} = 0$
\STATE $diag(D) = 0$
\ENSURE ~~\\ 
A distances matrix $D\in \mathbb{R}^{N\times N}$ with its elements simulates the distance between data points
\end{algorithmic}
\end{algorithm}

WDT is applied to the original affinity matrix $W$, and it transforms $W$ into a simulated distances matrix. After optimization for simulated distances, $D^*\in \mathbb{R}^{N\times N}$ is generated from $D$ and it needs transforming back into affinity matrix so that spectral clustering can be applied. Thus the inverse transformation of WDT which is called DWT is proposed as follows.

\begin{algorithm}
\caption{Distance-Weight Transformation (DWT)}
\label{alg:5}
\begin{algorithmic}[1]
\REQUIRE ~~\\ 
An optimized distance matrix $D^*\in \mathbb{R}^{N\times N}$
\STATE $w_{ij} = \tfrac{1}{d^*_{ij}}$ if $d^*_{ij} \not= \infty$\quad or\quad $w_{ij} = 0$ if $w_{ij} = \infty (i \not= j)$
\STATE $diag(W^*) = 0$
\ENSURE ~~\\ 
$W^*\in \mathbb{R}^{N\times N}$ which shows a more authentic relationship between data points
\end{algorithmic}
\end{algorithm}

\subsection{Optimization strategy for simulated distances matrix D}
After WDT, the weights are transformed into simulated distances so that graph-based optimization can be applied. $Zeros$ in $W$ means no similarity between data points, so the corresponding distance is defined as $infinity$. Elements in $D$ such as $d_{ij}$ stands for the distance from data point $x_i$ to $x_j$, and it’s obvious that $D$ is symmetric which has $zeros$ as diagonal elements. In the first stage, elements in $W$ fail to reveal the authentic relations between data points, thus the simulated distances generated from $W$ are not precise either. In the simulated undirected graph $G = (X, E)$ with vertex set $X = \left[ x_1, …,x_N \right]$, many potential connections are not established, which results in extremely overlarge distances or even infinite distances. So it’s a problem of how to minimize the distances between data points. We adopt the classic Floyd-Warshall algorithm [40] which aims to find the shortest paths between all pairs of vertices in a weighted graph with positive or negative edge weights, after which $D$ can be optimized as $D^*$ with shortest distances between all data points.

\begin{algorithm}
\caption{Floyd-Warshall}
\label{alg:6}
\begin{algorithmic}[1]
\REQUIRE ~~\\ 
$D\in \mathbb{R}^{N\times N}$ with simulated distances as elements, $N$ the number of rows in $D$
\STATE $D^{(0)}=D$
\STATE for $k = 1$ to $N$
\STATE \qquad let $D^{(k)} = (d^{(k)}_{ij})$ be a new $N\times N$ matrix
\STATE \qquad for $i = 1$ to $N$
\STATE \qquad\qquad for $j = 1$ to $N$
\STATE \qquad\qquad\qquad $d^{(k)}_{ij} =\ \operatorname{min}(d^{(k)}_{ij},\ d^{(k-1)}_{ik}+ d^{(k-1)}_{kj})$
\STATE $D^* = D^{(N)}$
\ENSURE ~~\\ 
$D^*$ with shortest distances as elements
\end{algorithmic}
\end{algorithm}

\paragraph{Theorem 1.} \emph{Floyd-Warshall minimizes the distances between high-dimensional data points.}
\begin{equation}
d^{*}_{ij} =\ \operatorname{min}(d_{ij},\ d_{ik} + d_{kj})
\end{equation}

\paragraph{Corollary 1.} \emph{3S-SC with GBTO optimizes the similarity between data points.}
\begin{equation}
\begin{split}
w^*_{ij}=\ \operatorname{DWT}(d^*_{ij}) =\ \operatorname{DWT}( \operatorname{min}(d_{ij}, d_{ik}+d_{kj}))
\\=\ \operatorname{DWT}( \operatorname{min}(\operatorname{WDT}(w_{ij}), \operatorname{WDT}(w_{ik})+\operatorname{WDT}(w_{kj})))
\\=\ \operatorname{max}(w_{ij},\ \operatorname{DWT}(\ \operatorname{WDT}(w_{ik})\ +\ \operatorname{WDT}(w_{kj})\ )\ )
\end{split}
\end{equation}
After traversal of all intermediate points $x_k$, $k\in$ $\{1,...,N \}$ , the Floyd–Warshall algorithm compares all potential paths through the graph $G$ between each pair of data points. The left upper of Figure 3 is the simulated graph $G = (X, E)$ as proposed in Section 3.1. After Floyd-Warshall, some new connections have been created which are shown as dotted lines in the right upper of Figure 3. Besides, some existing connections are optimized by minimizing the distances via intermediate nodes, such as the connection between node $x_i$ and $x_j$ which is optimized via intermediate node $x_k$.

\begin{figure}
  \begin{minipage}[t]{0.50\linewidth}
  \centering
  \includegraphics[height = 4cm]{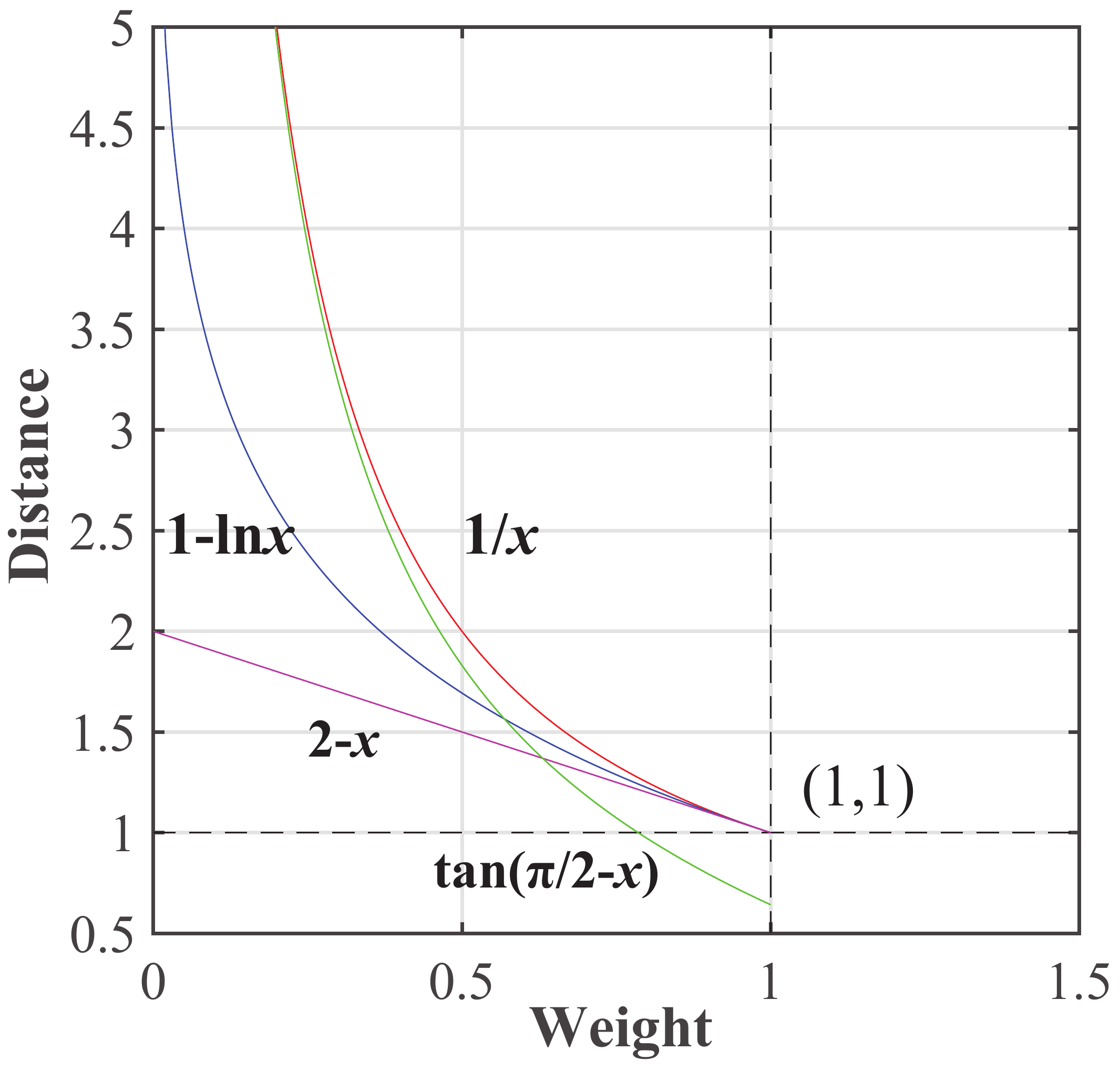}
  \caption{Comparison between transformations}
  \label{fig2}
  \end{minipage}%
  \begin{minipage}[t]{0.50\linewidth}
  \centering
  \includegraphics[height = 3cm]{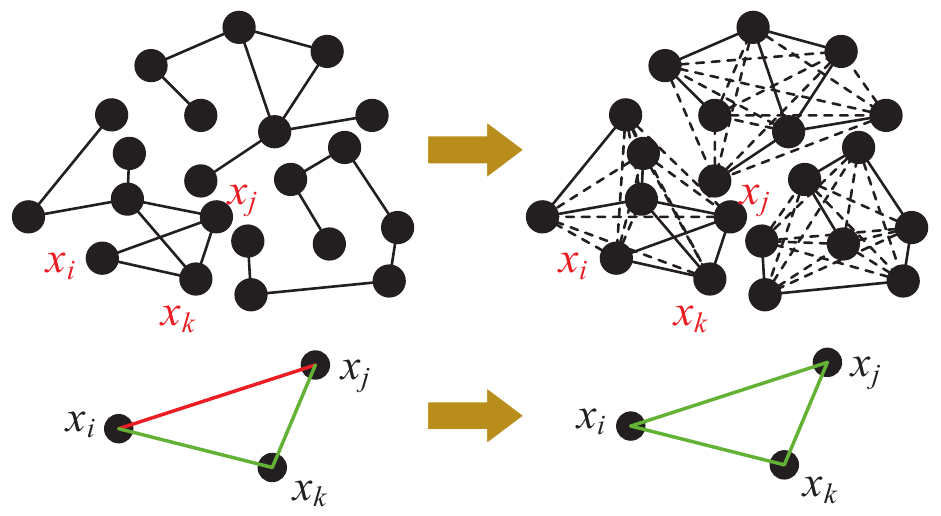}
  \caption{Changes in $G$ after GBTO}
  \label{fig3}
  \end{minipage}
\end{figure}

\subsection{3S-SC applied with GBTO}
With the implementation of GBTO, 3S-SC can be applied to most of spectral clustering based SC methods. 3S-SC with GBTO applied with Algorithm 5 is defined as Hard 3S-SC, which means both zero and nonzero elements in affinity matrix $W$ are optimized. This will lead to more nonzero elements in affinity matrix. For Soft 3S-SC, restriction as $d_{ij} \not= \infty$ is added in step 6 of Algorithm 5 to ensure only nonzero elements in $W$ are optimized, in which way only the existing relations are optimized and there is no risk that wrong connections between subspaces are established. Compared with Hard 3S-SC, Soft 3S-SC is a more conservative optimization method which improves spectral clustering accuracy without changing sparsity.
\begin{equation}
c_{i}=\underset{c_{i}}{\arg \min }\left\|c_{i}\right\|_{0} \text { s.t. } x_{i}=X c_{i}, c_{ii}=0
\end{equation}
In the first stage, different techniques are applied to solve the problem in (5). Usually it's simplified as a $\ell_1$ or $\ell_2$ problem which is not NP hard. $c_i$ in $C = [c_1, …,c_N]$ are coefficients for $x_i$ to be written as a linear combination of other points based on self-expressiveness. Multiple solutions for $C$ vary from each other, so self-expressiveness could not show the global linear correlation of data.

\paragraph{Lemma 1.} \emph{3S-SC with GBTO optimizes the expression of linear correlation of data points.}

With $S_i$ standing for subspace, (1) shows the self-expressiveness property of data lying in the same subspace, which can be derived that:
\begin{equation}
x_i\in S_i,\ \forall j = 1,...,N,\ c_{ij}\not= 0 \Longrightarrow x_i \in S_i.
\end{equation}
  However, the coefficients matrix $C$ obtained in the first stage doesn't possess good self-expressiveness. For example, if we initialize $x_i\in S_i, x_k\in S_k, x_j\in S_j$, in some special cases:
\begin{equation}
 c_{ij}=0,\ c_{ik}\not=0,\ c_{kj}\not=0 \Longrightarrow x_i \in S_k,\ x_j \in S_k,
\end{equation}
 where $x_i$ and $x_j$ belong to the same subspace, while the opposite conclusion can be derived from $c_{ij}=0$. More commonly, for some nonzero entries in $C$:
\begin{equation}
 c_{ij}\ll c_{ik},\ c_{ij}\ll c_{kj}\ \Longrightarrow d_{ij}\gg d_{ik},\ d_{ij}\gg d_{kj}\quad \mbox{or equivalently}\quad d_{ij} \gg d_{ik} + d_{kj},
\end{equation}
the distance from intermediate point $x_k$ to $x_i$ and $x_j$ are small while $x_i$ is far from $x_j$, which is irrational in ternary relationship. After GBTO, the distances between data points are minimized and the similarity is optimized, which can be viewed as optimization for $C$ as well. That is:
\begin{equation}
\begin{split}
c^*_{ij} = \tfrac{1}{2} w^*_{ij} =\ \tfrac{1}{2}\operatorname{max}(w_{ij},\ \operatorname{DWT}(\ \operatorname{WDT}(w_{ij})\ +\ \operatorname{WDT}(w_{kj})\ )\ )
\\=\ \operatorname{max}(\tfrac{1}{2}w_{ij},\ \operatorname{DWT}(\ \operatorname{WDT}(\tfrac{1}{2}w_{ij})\ +\ \operatorname{WDT}(\tfrac{1}{2}w_{kj})\ )\ )
\\=\ \operatorname{max}(c_{ij},\ \operatorname{DWT}(\ \operatorname{WDT}(c_{ij})\ +\ \operatorname{WDT}(c_{kj})\ )\ )
\end{split}
\end{equation}
Thus, 3S-SC with GBTO optimizes the data representation to possess better linear correlation expression.

\paragraph{Lemma 2.} \emph{In spectral clustering, higher similarity of edges inside subgraphs leads to smaller cut.}

State-of-the-art spectral clustering adopt Ncut [34] to gain a more accurate result.Ncut is defined as :
\begin{equation}
Ncut(A, B)=\frac{\operatorname{cut}(A, B)}{\operatorname{assoc}(A, G)}+\frac{\operatorname{cut}(B, A)}{\operatorname{assoc}(B, G)},
\end{equation}
where $cut$ is the degree of dissimilarity between these two subgraphs $A$ and $B$, which can be computed as total weight of the edges that have been removed, and $\operatorname{assoc}(A, G)=\sum_{u \in A, t \in G} w(u, t)$is the total connection from vertexes in $A$ to all vertexes in graph $G$. It can be rewritten as:
\begin{equation}
Ncut(A, B)=\frac{\sum_{\left(e_{i}>0, e_{j}<0\right)}-w_{i j} e_{i} e_{j}}{\sum_{e_{i}>0} d_{i}}+\frac{\sum_{\left(e_{i}<0, e_{j}>0\right)}-w_{i j} e_{i} e_{j}}{\sum_{e_{i}<0} d_{i}},
\end{equation}
where $e_i = 1$ if vertex $x_i \in A$ and $e_i-1$ if $x_i \notin A$, and $d(i) = \sum_{j} w_{ij}$. For the task of partitioning graph into $k$ pieces, the Ncut for SC can be rewritten as:
\begin{equation}
Ncut\left(A_{1}, A_{2}, \ldots A_{k}\right)=\frac{1}{2} \sum_{i=1}^{k} \frac{\operatorname{cut}\left(A_{i}, \overline{A}_{i}\right)}{W\left(A_{i}\right)} = \sum_{i=1}^{k} \frac{\operatorname{cut}\left(A_{i}, \overline{A}_{i}\right)}{\operatorname{vol}\left(A_{i}\right)},
\end{equation}
where $\operatorname{vol}\left(A_{i}\right)= \sum_{j \in A_{i}} d_{j}$ is defined as the sum of the weights of all edges in subgraph $A_i$ whose nodes belong to $S_i$, and $\overline{A}_{i}$ is graph without $A_i$. After applied with 3S-SC, $A^*_i$ with optimized similarity is generated. The increment of weights inside subgraphs is much larger than that between subgraphs especially in graph with more subgraphs and more nodes per subgraph. That is:
\begin{equation}
\operatorname{vol}(A^*_i)- \operatorname{vol}(A_i) \gg  \operatorname{cut}\left(A^*_{i}, \overline{A_{i}^*}\right)-\operatorname{cut}\left(A_{i}, \overline{A}_{i}\right).
\end{equation}
This is obvious since data points belong to the same subspace have larger similarity and more connections inside subgraphs are established. So it can be derived that:
\begin{equation}
\sum_{i=1}^{k} \frac{\operatorname{cut}\left(A^*_{i}, \overline{A^*_{i}}\right)}{\operatorname{vol}\left(A^*_{i}\right)} < \sum_{i=1}^{k} \frac{\operatorname{cut}\left(A_{i}, \overline{A}_{i}\right)}{\operatorname{vol}\left(A_{i}\right)}
\end{equation}
\paragraph{Theorem 2.} \emph{3S-SC with GBTO improves accuracy and connectivity by higher similarity which is closer to the truth.}
\begin{equation}
\operatorname{sparsity}(D^*) \leq \operatorname{sparsity}(D) \Longrightarrow
\operatorname{sparsity}(W^*) \leq \operatorname{sparsity}(W).
\end{equation}
This can be concluded from (4) and (15), which show how 3S-SC affects the similarity. This means 3S-SC with GBTO improves clustering accuracy and connectivity by optimized similarity.


\section{Experiments}
We compare the performance of spectral clustering based SC methods with these implemented as 3S-SC methods, including SSC-$\ell_1$, OMP, ENSC, LRR, LSR and BDR, which contain $\ell_0$, $\ell_1$, $\ell_2$ and nuclear norm. These methods applied with hard 3S-SC are prefixed with ’3S’, such as 3S-SSC-$\ell_1$. Those applied with soft 3S-SC are prefixed with 'soft'. Parameters are set as recommended ($\lambda=0.9$ for ENSC, $\lambda=50, \gamma=1$ for BDR). Experiments are set on synthetic data, handwritten digits set MNIST [35] and USPS [36], and human faces set Extended Yale B (EYaleB)[37] with 50 trails. Connectivity, Clustering Accuracy and Normalized Mutual Information (NMI) [38][39] are metrics to evaluate the performance. Connectivity is defined as the second smallest eigenvalue $\lambda_2$ of the normalized Laplacian $L=I-D^{-1/2}WD^{-1/2}$, where $D = Diag(W\cdot 1)$ is the degree matrix of graph $G$. NMI quantifies the amount of information obtained by clustering results compared with the ground-truth.

\begin{figure}[h]
  \centering
  \subfigure{
  \includegraphics[height = 3.1cm]{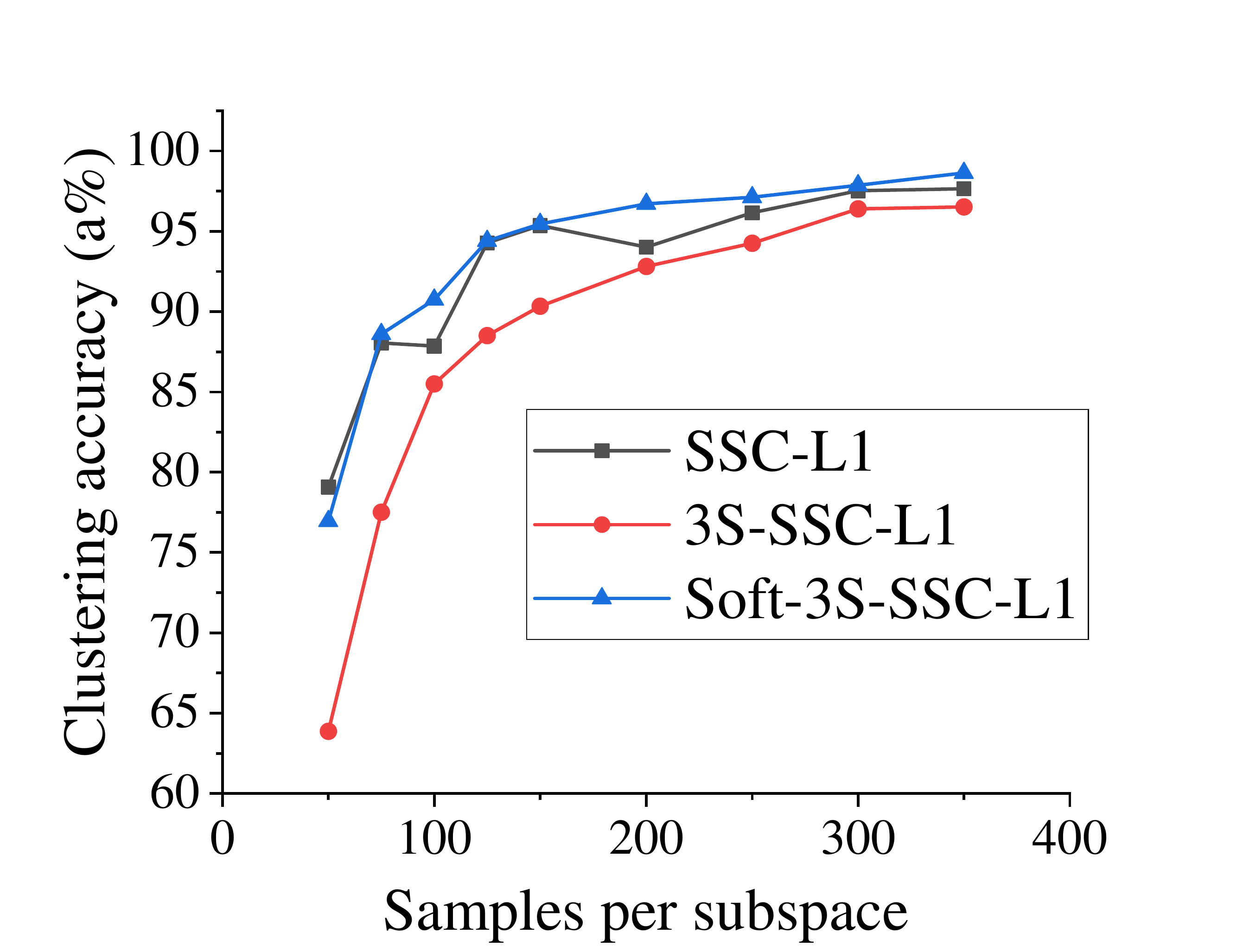}}
  \subfigure{
  \includegraphics[height = 3.1cm]{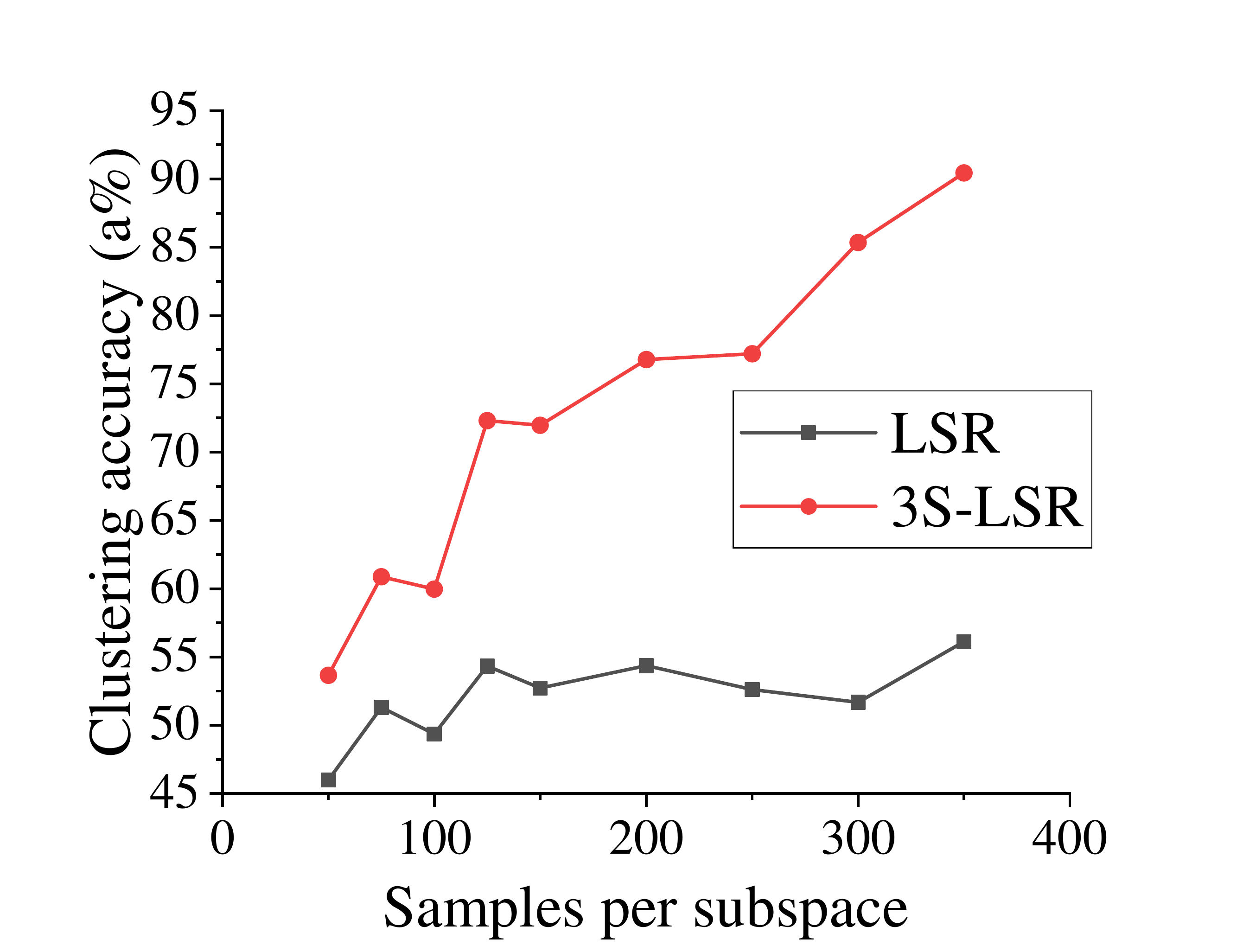}}
  \subfigure{
  \includegraphics[height = 3.1cm]{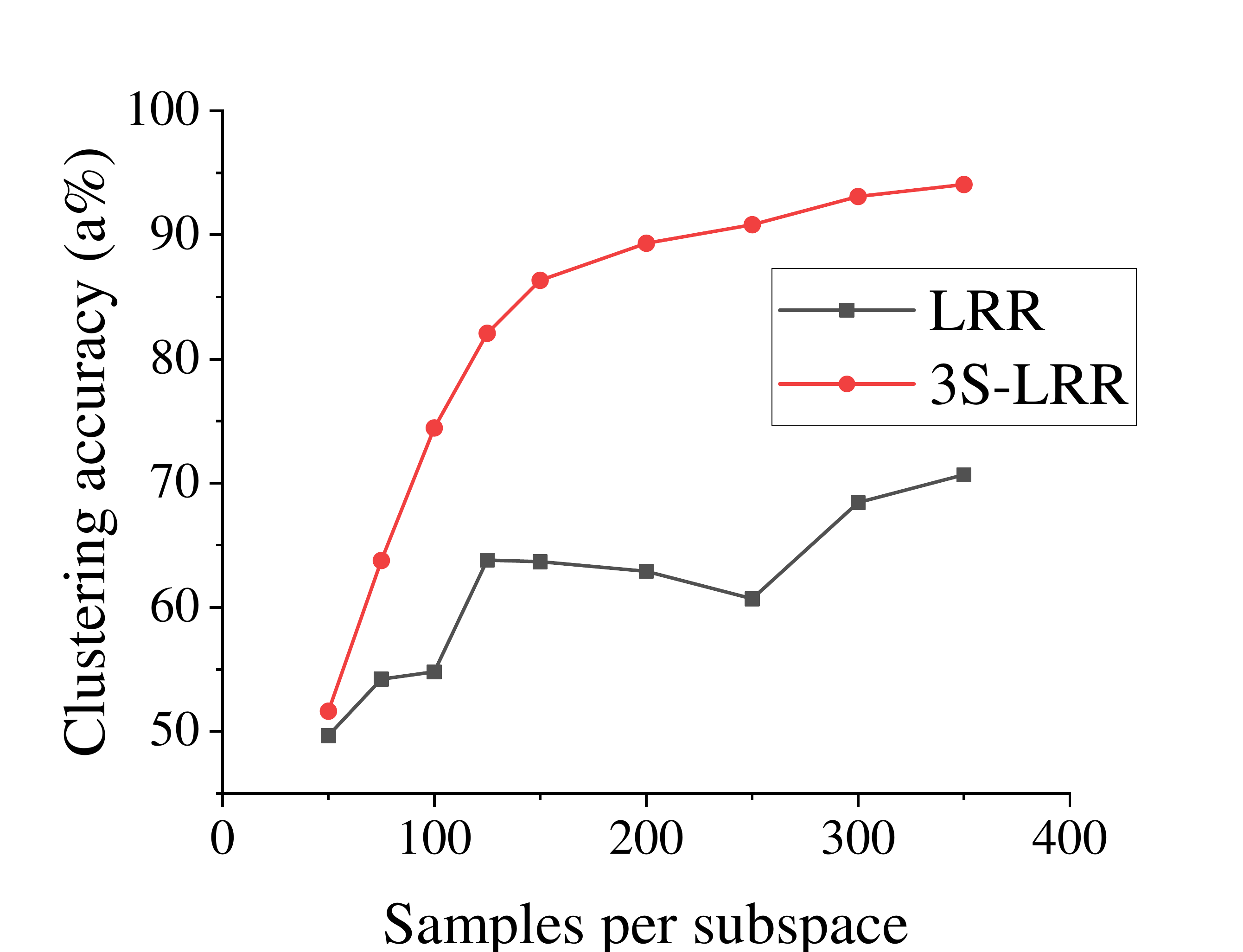}}
  \subfigure{
  \includegraphics[height = 3.1cm]{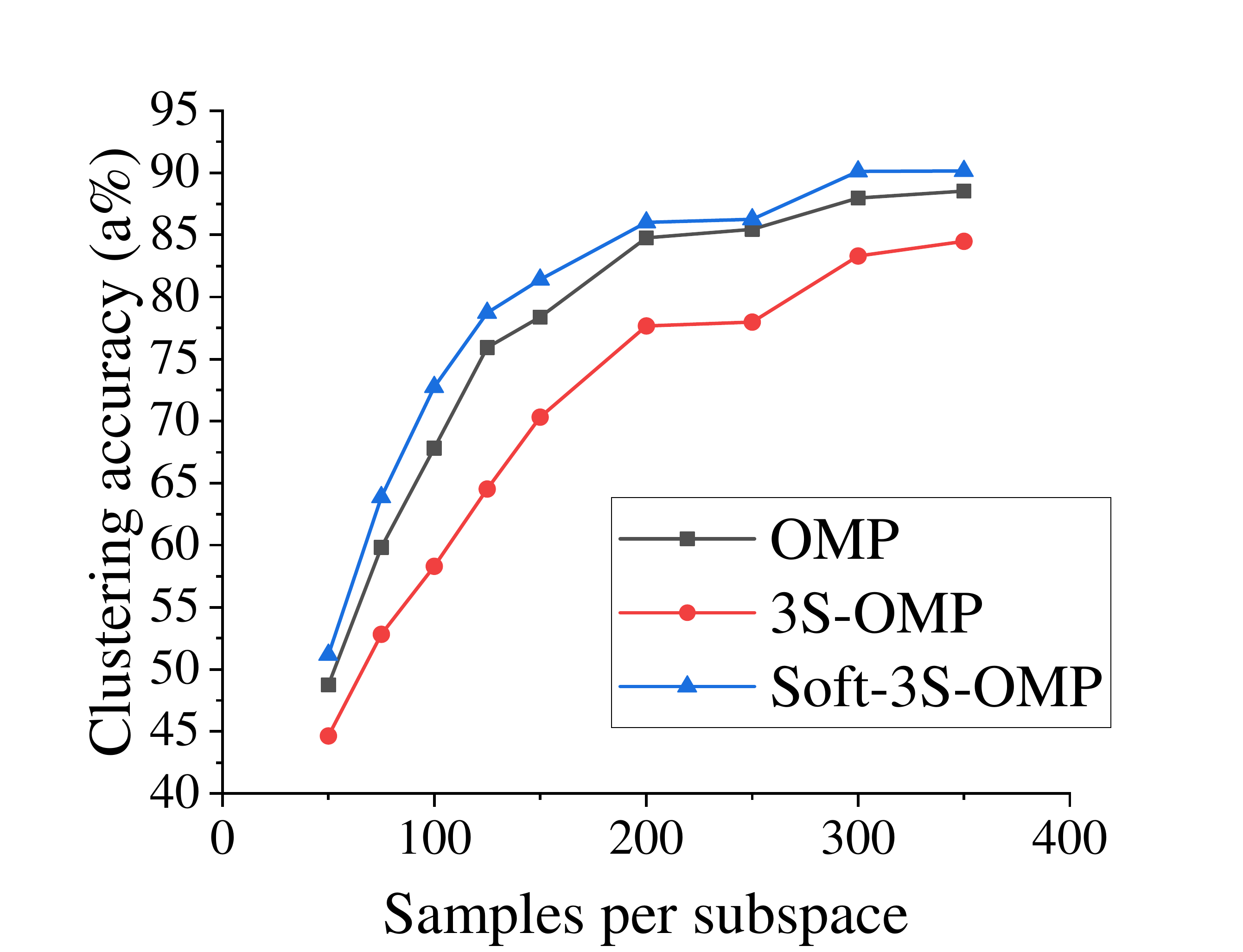}}
  \subfigure{
  \includegraphics[height = 3.1cm]{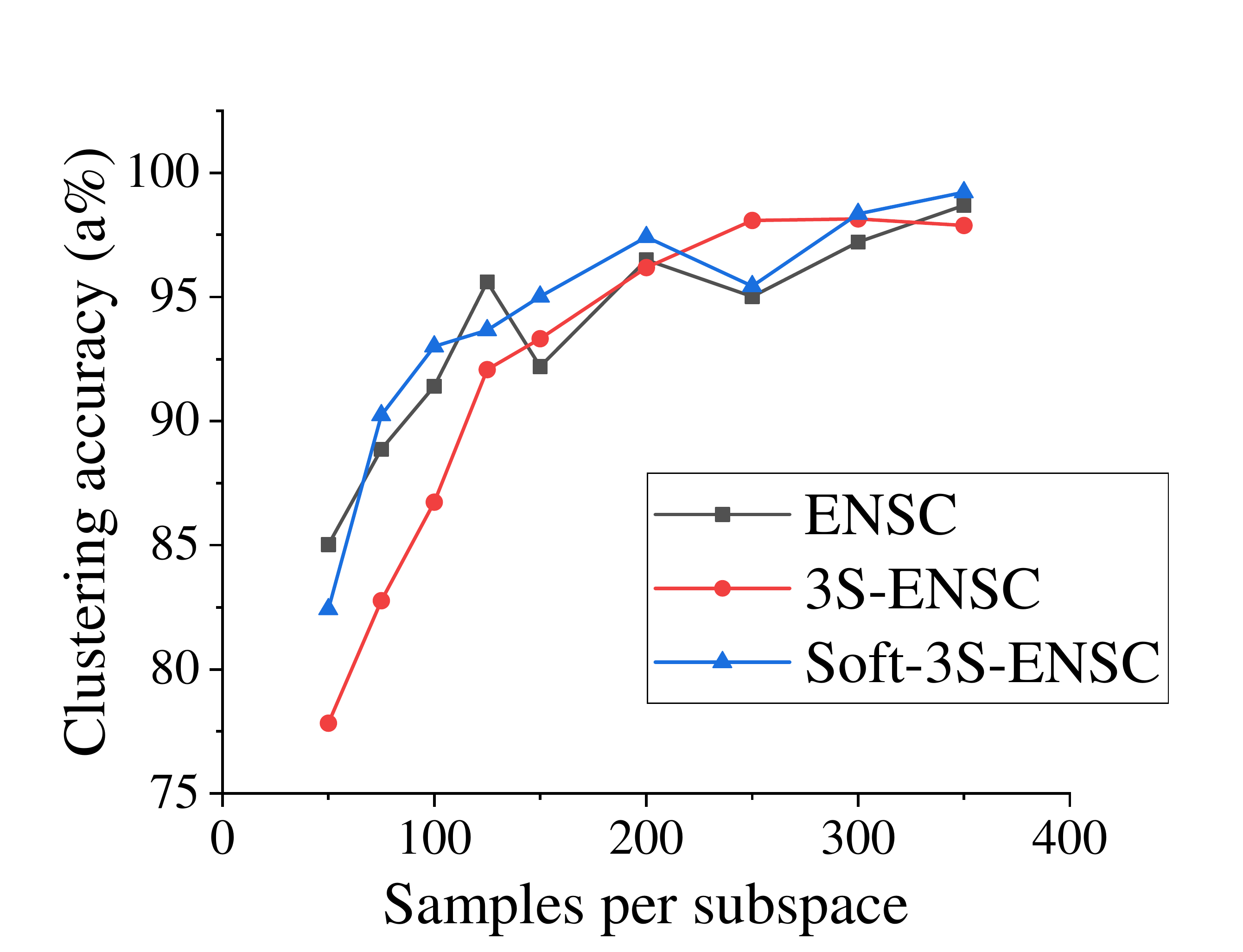}}
  \subfigure{
  \includegraphics[height = 3.1cm]{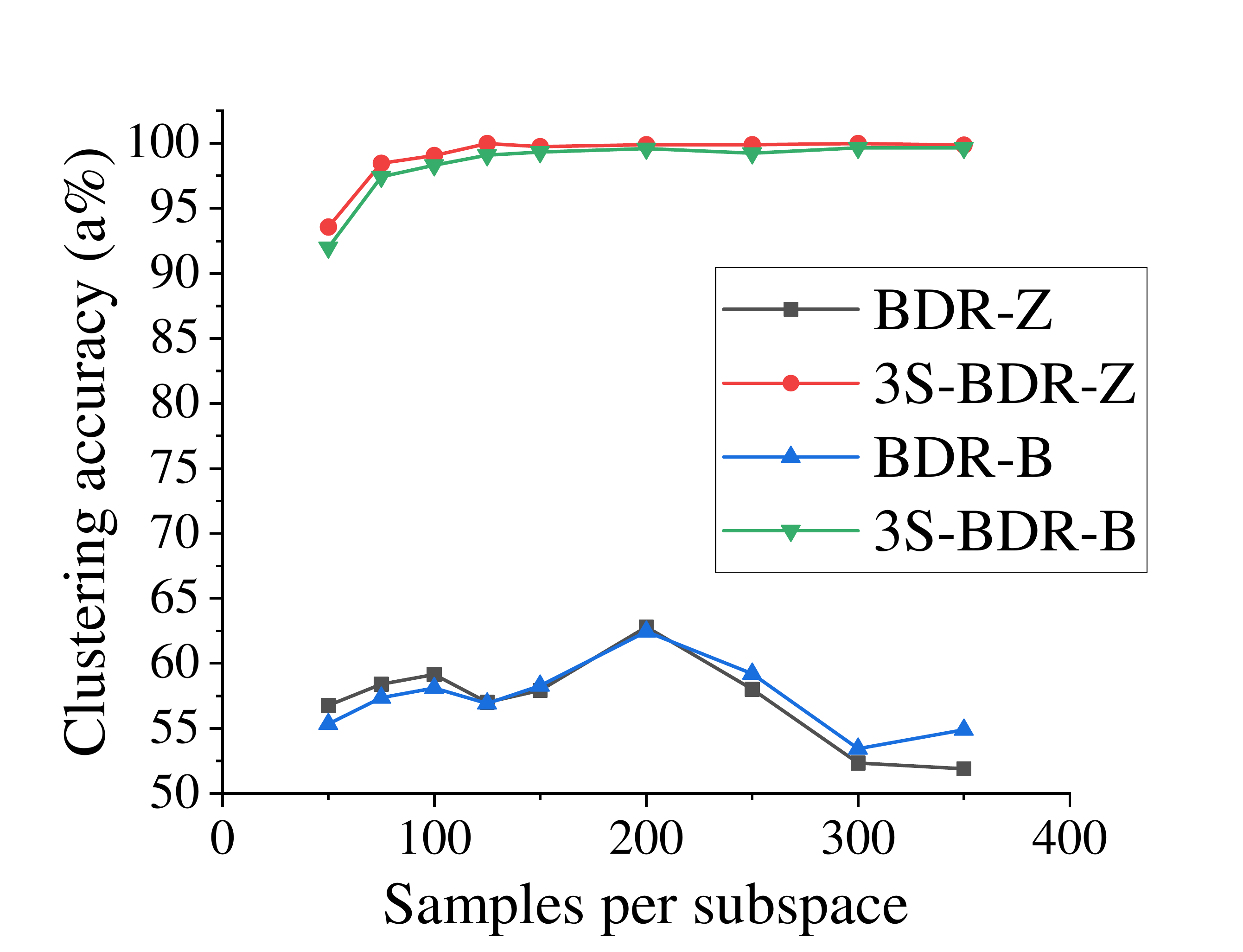}}
  \caption{Accuracy comparison on synthetic data}
\end{figure}

\begin{figure}[h]
  \centering
  \subfigure[NMI]{
  \includegraphics[height = 3.2cm]{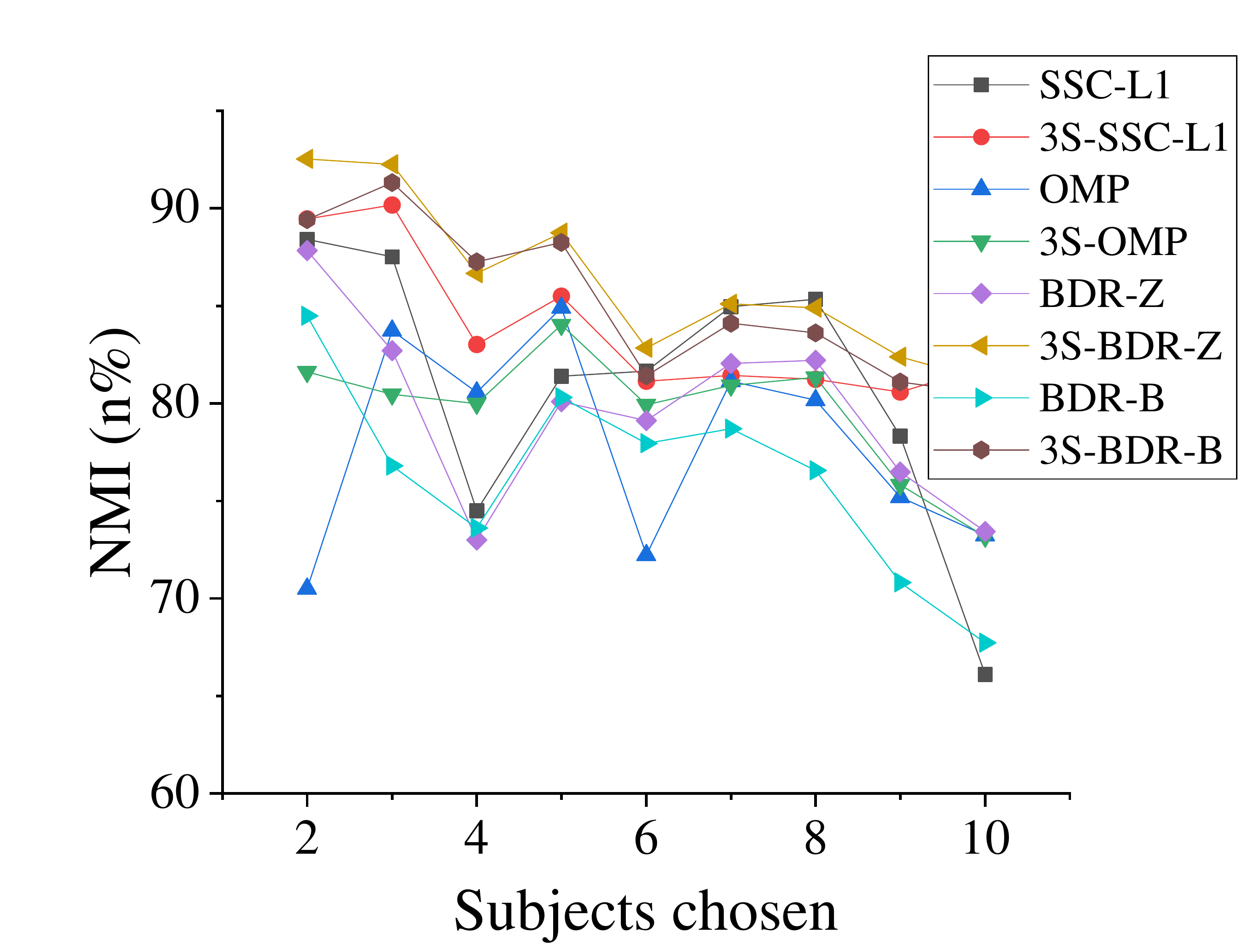}}
  \subfigure[NMI]{
  \includegraphics[height = 3.2cm]{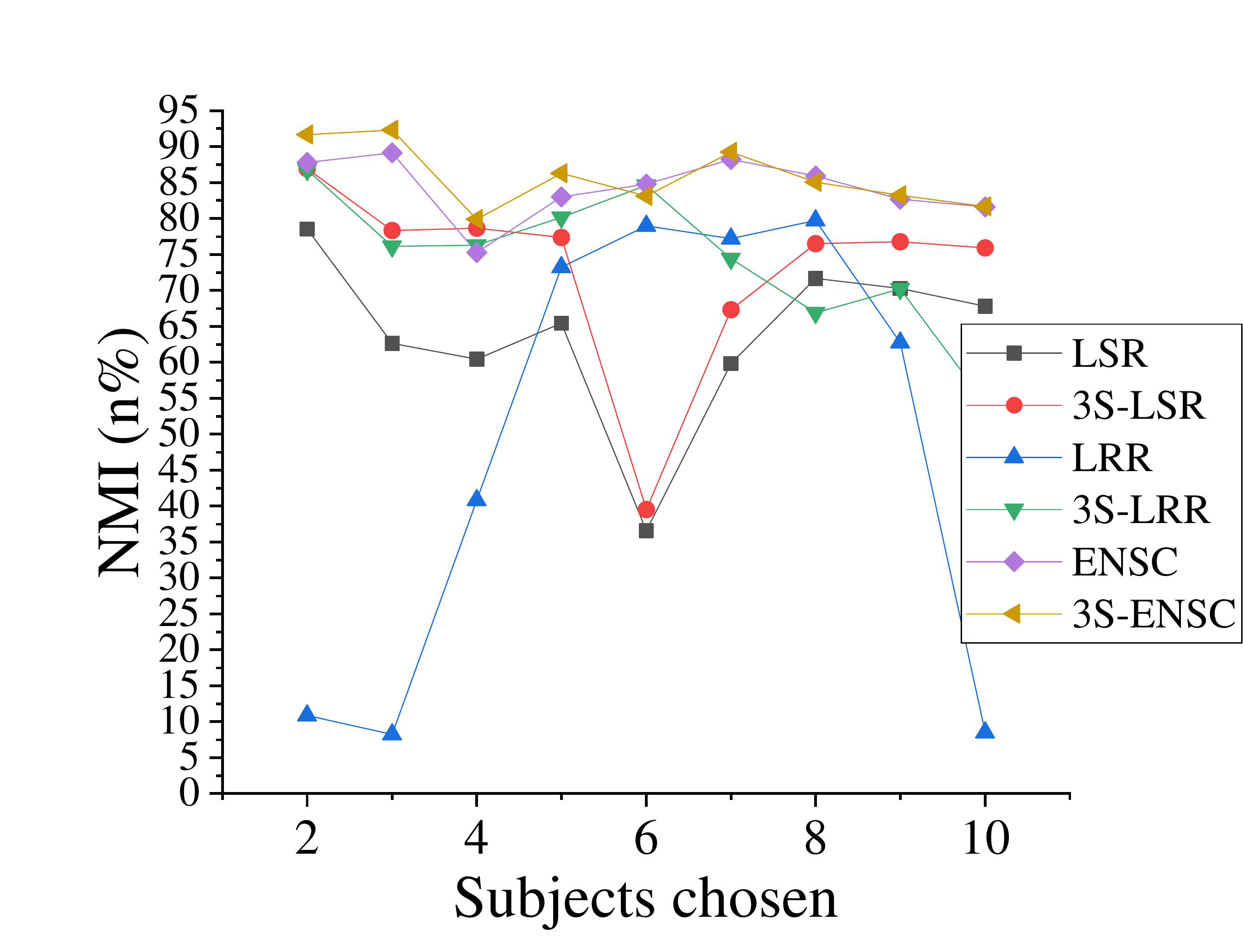}}
  \subfigure[Connectivity]{
  \includegraphics[height = 3.2cm]{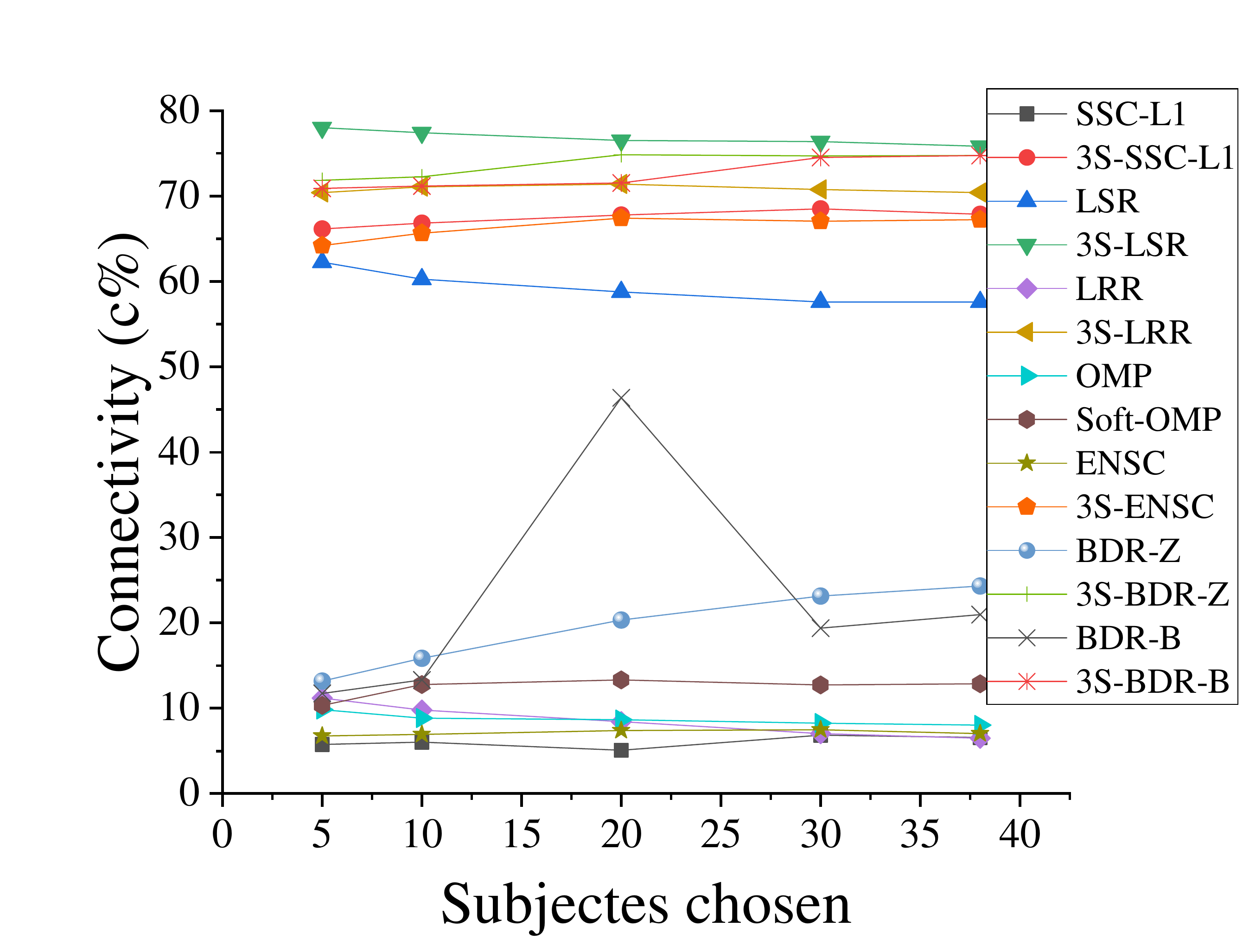}}
  \caption{Both (a) and (b) show NMI on MNIST, (c) shows Connectivity on EYaleB. }
\end{figure}

\textbf{Synthetic Experiments.} Figure 4 shows how clustering accuracy changes with samples per subspace on synthetic data with $n = 5$ subspaces each of dimension $d = 8$ in ambient space of dimension $D = 12$ ($D=18,d=7$ for BDR). Hard 3S-SC improves accuracy for LSR, LRR and BDR, while soft 3S-SC works well in SSC-$\ell_1$, OMP and ENSC with $\ell_0$ and $\ell_1$ norm. NMI synchronously changes with accuracy, and connectivity of 3S-SC methods is usually higher than 80\%.

\textbf{Clustering Handwritten Digits.} Experiments on two data sets are set differently. The number of subjects chosen to be clustered changes on MNIST, while samples per subspace changes on USPS. The feature vectors for images in MNIST are projected to dimension 500 and 200 for USPS by PCA. As is shown in Table 1, Figure 5(a) and 5(b), in real-world data sets of handwritten digits, 3S-SC methods usually obtain higher accuracy and NMI. In terms of data sets with more samples per subspace, GBTO works better due to more authentic relations between data points.

\begin{minipage}{\textwidth}
 \begin{minipage}[t]{0.5\textwidth}\small
  \centering
     \makeatletter\def\@captype{table}\makeatother\caption{Accuracy(\%) on USPS}

    \begin{tabular}{lrrrrr}
    \toprule
    Samples     & 50    & 100   & 200   & 400\\   
    \midrule
    SSC-$\ell_1$ & 62.68 & 65.05 & 60.97 & 60.13\\
    3S-SSC-$\ell_1$ & \textbf{70.50}  & \textbf{78.45} & \textbf{73.13} & \textbf{82.30}  \\
    \midrule
    LSR   & 36.10  & 68.18 & 71.09 & 71.11\\
    3S-LSR & \textbf{37.24} & \textbf{71.80}  & \textbf{73.79} & \textbf{75.88} \\
    \midrule
    LRR   & 67.42 & 64.86 & 61.86 & 62.85 \\
    3S-LRR & \textbf{71.01} & \textbf{70.25} & \textbf{69.02} & \textbf{71.74} \\
    \midrule
    OMP   & 58.90  & 61.39 & 59.99 & 61.62 \\
    3S-OMP & \textbf{67.28} & \textbf{70.57} & \textbf{74.41} & \textbf{73.13} \\
    \midrule
    ENSC  & 57.94 & 60.34 & 60.62 & 59.21 \\
    3S-ENSC & \textbf{63.50}  & \textbf{73.87} & \textbf{69.09} & \textbf{71.48} \\
    \midrule
    BDR-Z & 68.86 & 66.45 & 60.07 & 53.54 \\
    3S-BDR-Z & \textbf{71.60}  & \textbf{72.16} & \textbf{73.28} & \textbf{72.55}\\
    \midrule
    BDR-B & 68.18 & 65.98 & 64.61 & 57.35 \\
    3S-BDR-B & \textbf{70.41} & \textbf{72.50}  & \textbf{73.53} & \textbf{72.99} \\
    \bottomrule
    \end{tabular}%
  \end{minipage}
  \begin{minipage}[t]{0.5\textwidth}\small
   \centering
        \makeatletter\def\@captype{table}\makeatother\caption{Accuracy(\%) on Extended Yale B}
    \begin{tabular}{lrrrrr}
    \toprule
     Subjects   & 5     & 10    & 20    & 30 \\   
    \midrule
    SSC-$\ell_1$ & 76.28 & 54.67 & 54.19 & 59.06 \\ 
    3S-SSC-$\ell_1$ & \textbf{86.09} & \textbf{65.87} & \textbf{70.51} & \textbf{78.60}\\  
    \midrule
    LSR   & 80.31 & \textbf{71.46} & 57.96 & 57.43\\ 
    3S-LSR & \textbf{93.55} & 67.03 & \textbf{65.39} & \textbf{65.47}\\ 
    \midrule
    LRR   & 70.87 & 61.59 & 54.73 & 53.11\\ 
    3S-LRR & \textbf{79.99} & \textbf{66.76} & \textbf{57.67} & \textbf{56.15}\\ 
    \midrule
    OMP   & 96.60  & 87.89 & 81.34 & 78.77\\ 
    Soft-OMP & \textbf{97.28} & \textbf{90.32} & \textbf{85.34} & \textbf{80.80}\\  
    \midrule
    ENSC  & \textbf{75.57} & 67.91 & 66.18 & 63.99\\ 
    3S-ENSC & 63.82 & \textbf{69.89} & \textbf{73.85} & \textbf{71.66}\\ 
    \midrule
    BDR-Z & \textbf{81.18} & 61.20  & 63.63 & 60.72\\ 
    3S-BDR-Z & 80.32 & \textbf{63.72} & \textbf{74.86} & \textbf{75.68}\\ 
    \midrule
    BDR-B & \textbf{87.67} & \textbf{72.35} & 69.26 & 66.42\\ 
    3S-BDR-B & 82.99 & 65.93 & \textbf{74.94} & \textbf{75.70}\\  
    \bottomrule
    \end{tabular}%
   \end{minipage}
\end{minipage}

\textbf{Clustering Human Faces with Varying Lighting.} Data points are images downsampled from $192\times 168$ to $48\times 42$. Table 2 shows accuracy and Figure 5(c) reports connectivity. For OMP, slight improvements has been made when applied with soft 3S-SC since no new connections are created. For ENSC and BDR, accuracy decrease of 3S-SC methods occurs when clustering 5 or 10 subjects because more connections between subspaces are established compared with inner connections when data sets are small. Considerable increase can be observed when clustered subjects are more than 10.
\begin{figure}[h]
  \centering
  \subfigure{
  \includegraphics[height = 3cm]{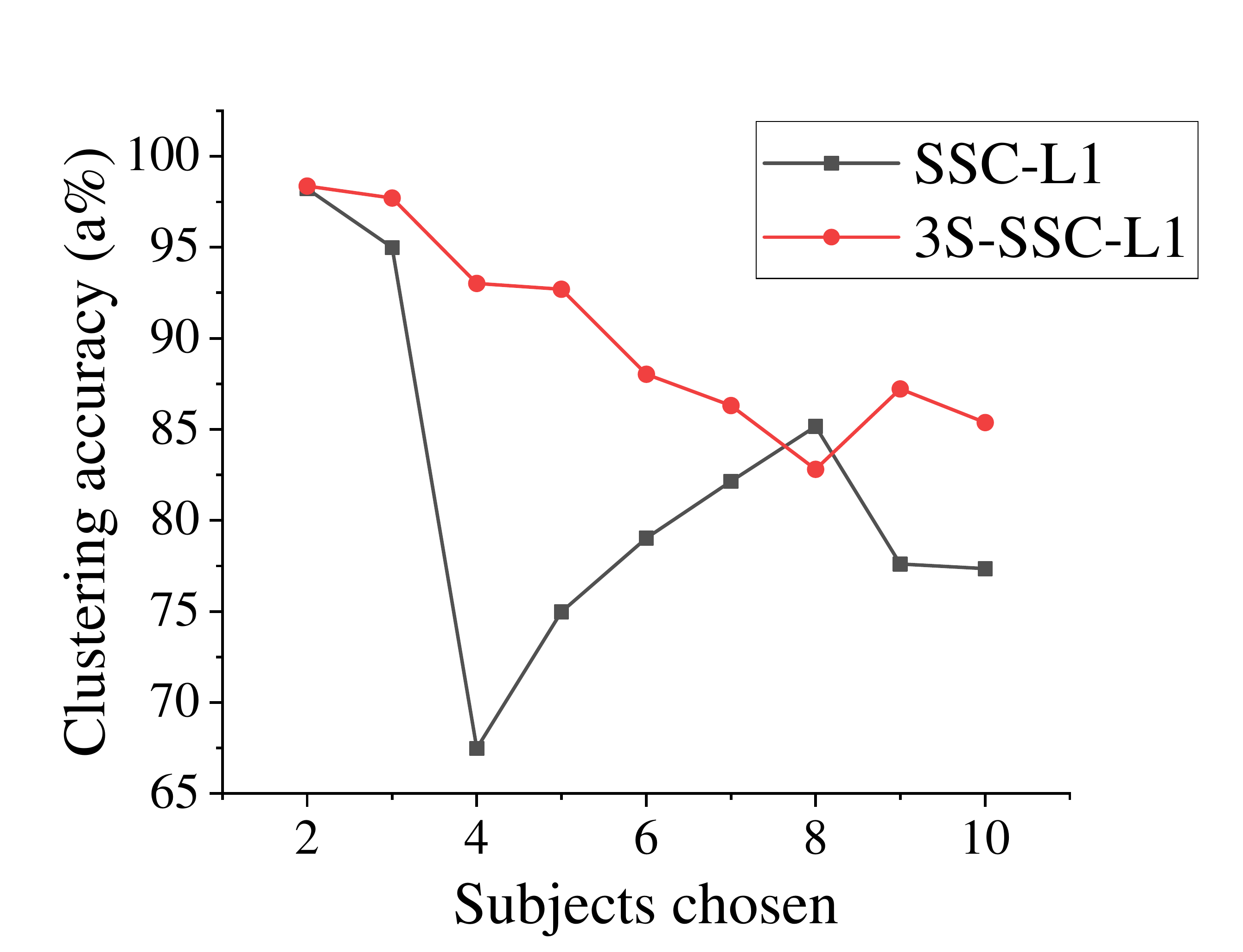}}
  \subfigure{
  \includegraphics[height = 3cm]{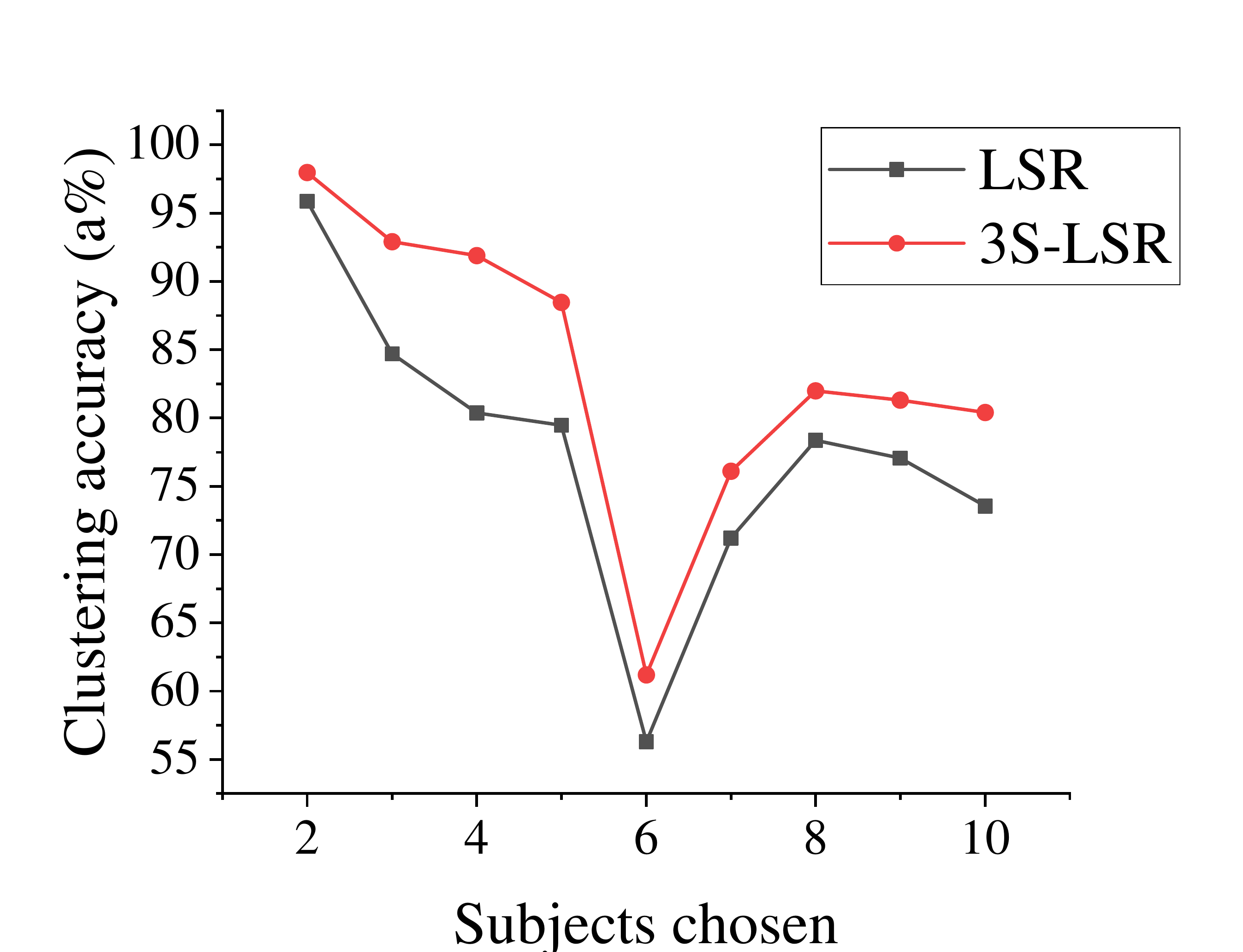}}
  \subfigure{
  \includegraphics[height = 3cm]{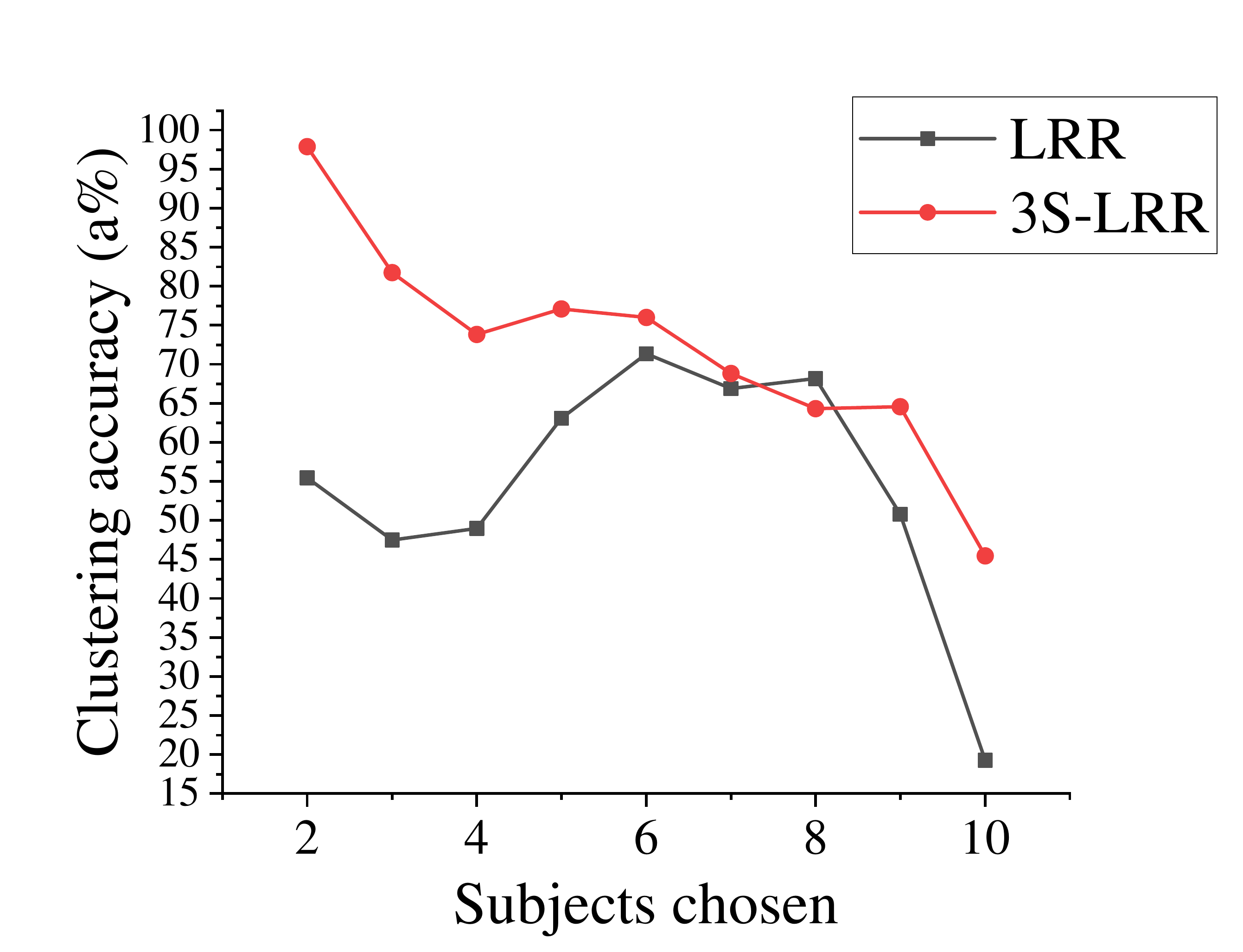}}
  \subfigure{
  \includegraphics[height = 3cm]{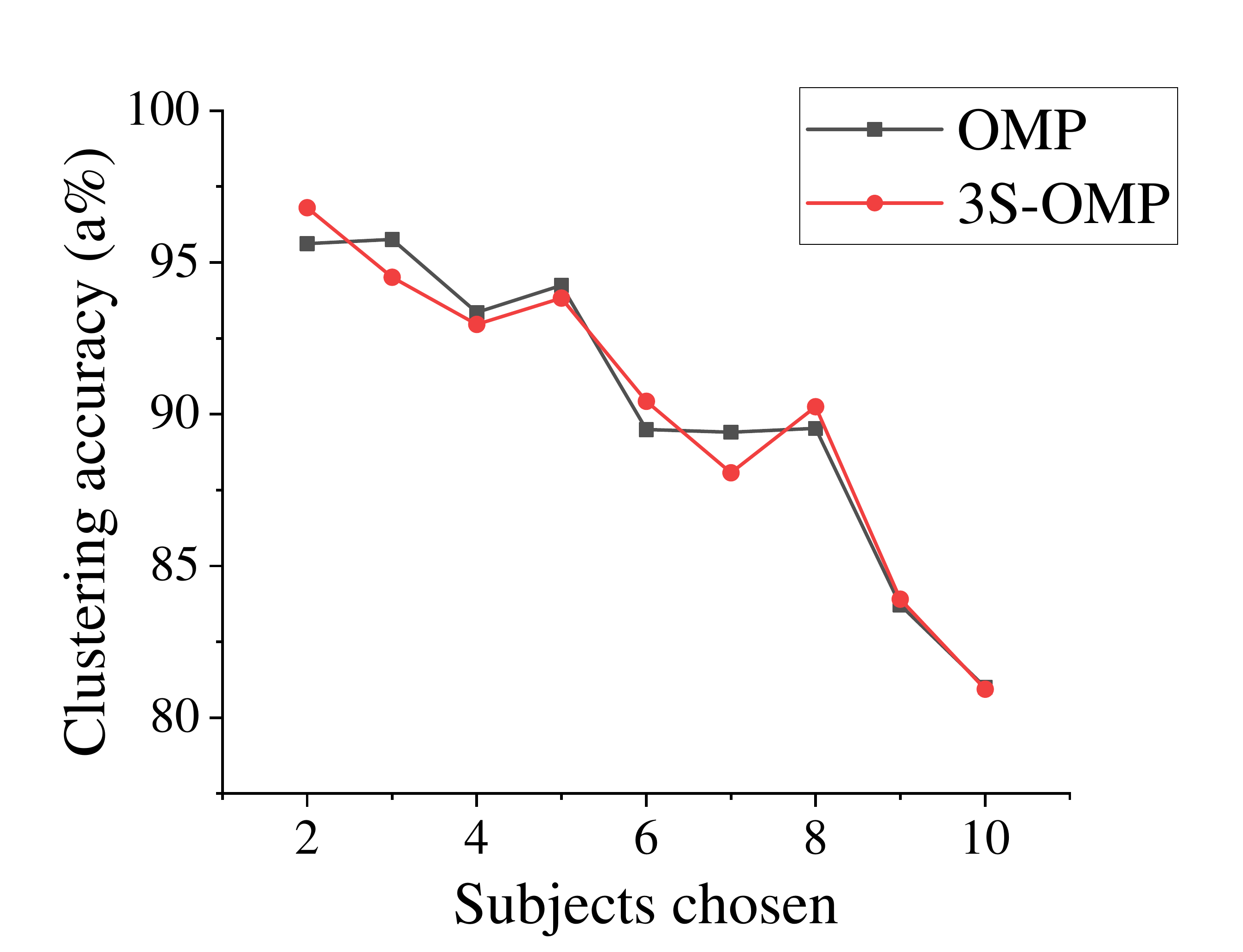}}
  \subfigure{
  \includegraphics[height = 3cm]{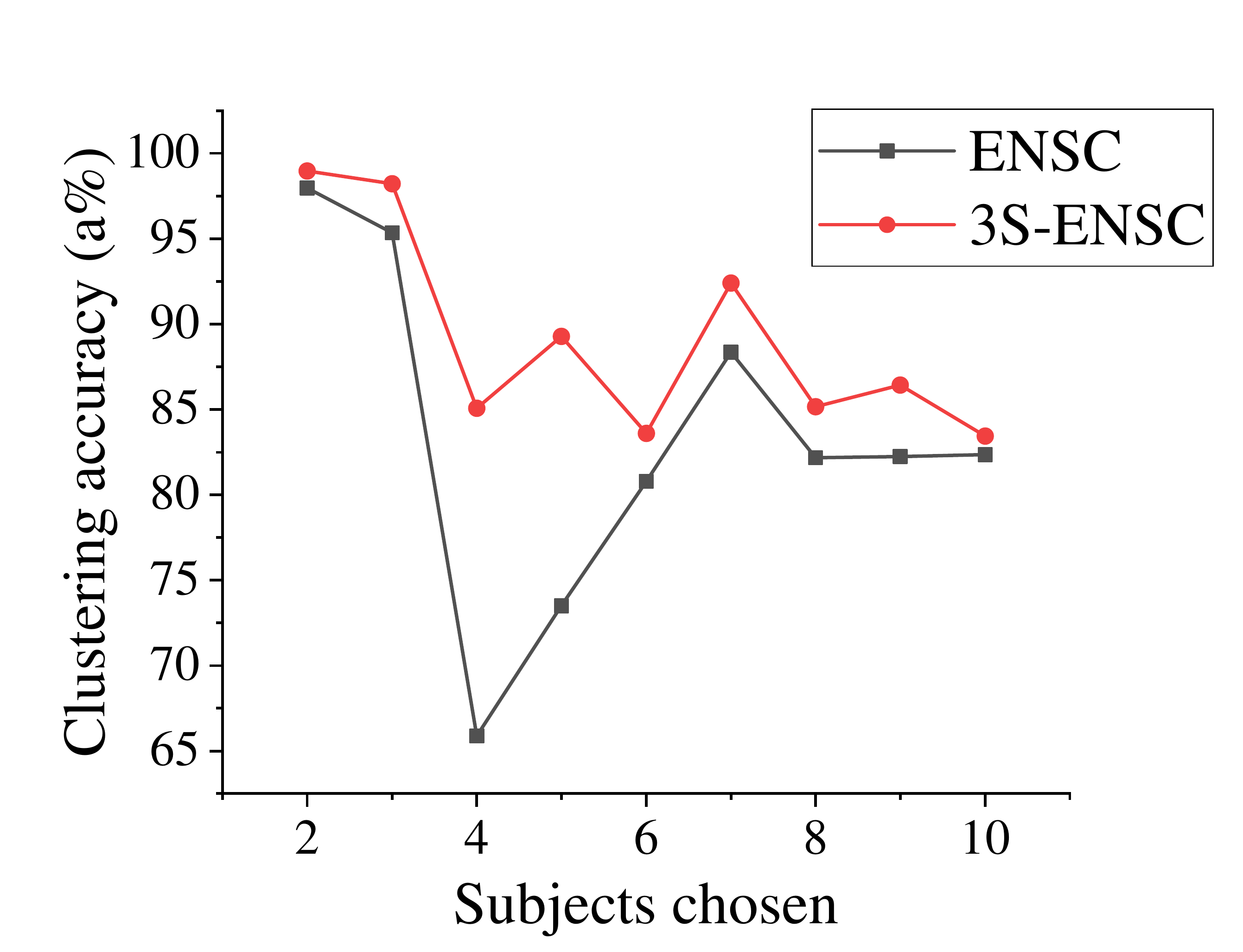}}
  \subfigure{
  \includegraphics[height = 3cm]{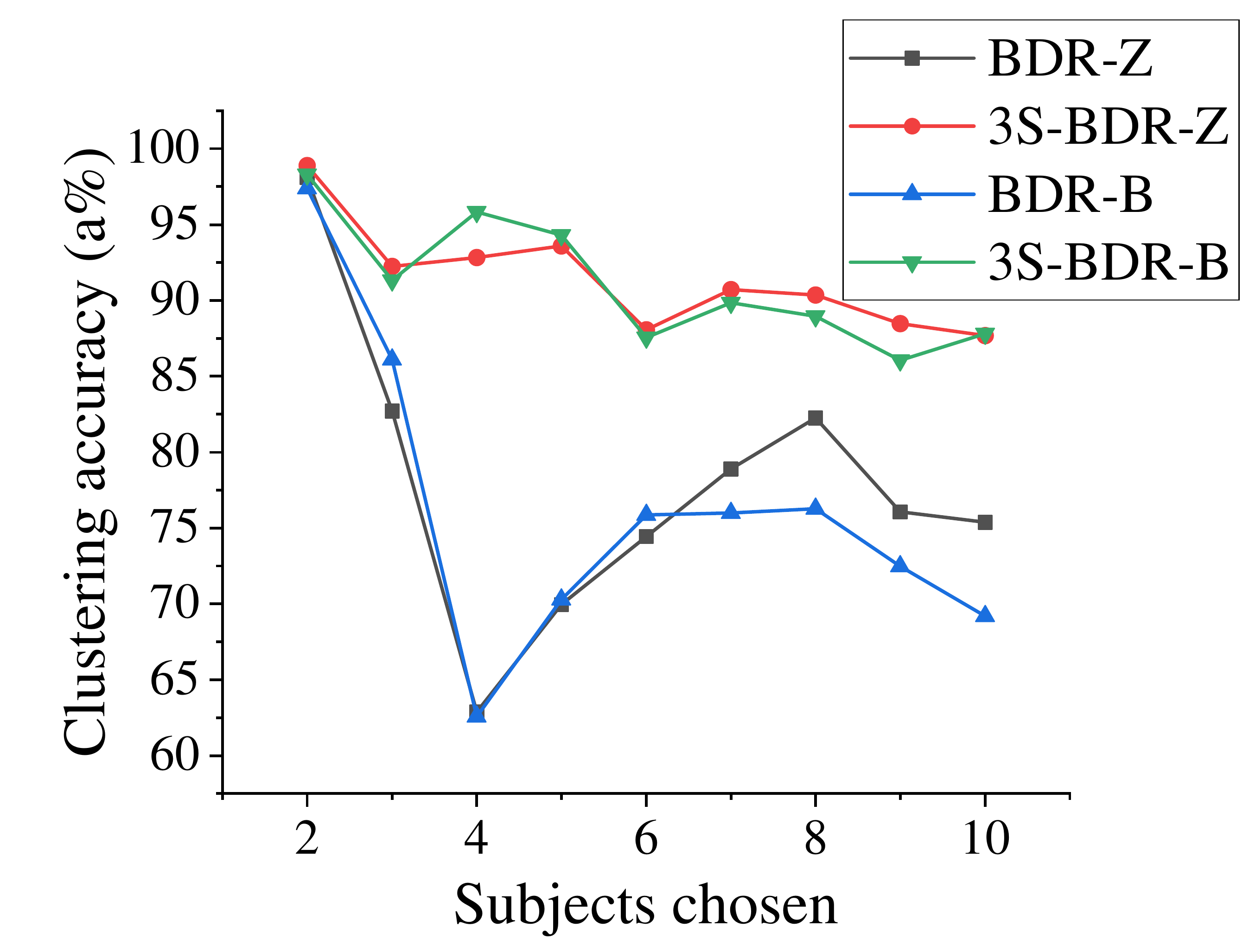}}
  \caption{Accuracy comparison on MNIST}
\end{figure}

The time complexity of Floyd-Warshall is $O\left(n^3 \right)$, while the time cost of 3S-SC is still acceptable even on limited computing resources. For larger and more complex data sets, techniques like vectorization for loops in ®MATLAB could reduce time. Moreover, the performance of BDR varies greatly when $\lambda$ and $\gamma$ are set differently, while 3S-BDR overcomes the over-dependence on parameters.

\section{Conclusion}
We proposed a Three-Stage Subspace Clustering framework (3S-SC) with two implementations (Hard \& Soft), in which Graph-Based Transformation and Optimization (GBTO) is applied to optimize the representation of authentic data distribution. 3S-SC is universal for SC methods with different regularizations, and the effectiveness of it is demonstrated on several data sets. We note that 3S-SC sometimes doesn’t work well in small data sets and this is left for future research.

\subsubsection*{Acknowledgments}
This work was supported in part by the Shenzhen Municipal Development and Reform Commission (Disciplinary Development Program for Data Science and Intelligent Computing), in part by Shenzhen International cooperative research projects GJHZ20170313150021171, and in part by NSFC-Shenzhen Robot Jointed Founding (U1613215).

\section*{References}

[1]	R. Basri and D. W. Jacobs, "Lambertian reflectance and linear subspaces,"  {\it IEEE Transactions on Pattern Analysis and Machine Intelligence}, vol. 25, no. 2, pp. 218-233, Feb. 2003.

[2]	W. Hong, J. Wright, K. Huang and Y. Ma, "Multiscale Hybrid Linear Models for Lossy Image Representation,"  {\it IEEE Transactions on Image Processing}, vol. 15, no. 12, pp. 3655-3671, Dec. 2006.

[3]	J. Costeira and T. Kanade, “A Multibody Factorization Method for Independently Moving Objects,” {\it International Journal of Computer Vision}, vol. 29, no. 3, pp. 159-179, 1998.

[4]	K. Kanatani, "Motion segmentation by subspace separation and model selection," in {\it Proceedings of IEEE 8th International Conference on Computer Vision}, vol.2, pp. 586-591, 2001

[5]	R. Vidal, “Subspace Clustering,” {\it Signal Processing Magazine}, vol. 28, no. 2, pp. 52-68, 2011.

[6]	R. Vidal, Y. Ma, and S. Sastry, “Generalized principal component analysis (GPCA),” {\it IEEE Transactions on Pattern Analysis and Machine Intelligence}, vol. 27, no. 12, pp. 1945–1959, Dec. 2005.

[7]	P. S. Bradley and O. L. Mangasarian, “k-plane clustering,” {\it Journal of Global Optimization}, vol. 16, no. 1, pp. 23–32, 2000.

[8]	P. Tseng, “Nearest q-flat to m points,” {\it Journal of Optimization Theory and Applications}, vol. 105, no. 1, pp. 249–252, 2000.

[9]	T. Zhang, A. Szlam, and G. Lerman, “Median k-flats for hybrid linear modeling with many outliers,” in {\it Proceedings of IEEE 12th International Conference on Computer Vision Workshops}, pp. 234–241, 2009.

[10]	P. Agarwal and N. Mustafa, “k-means projective clustering,” in {\it Proceedings of ACM Symposium on Principles of Database Systems}, pp. 155–165, 2004.

[11]	T. E. Boult and L. G. Brown, “Factorization-based segmentation of motions,” in {\it Proceedings of IEEE Workshop Motion Understanding}, pp. 179–186, 1991.

[12]	Leonardis, H. Bischof, and J.Maver, “Multiple eigenspaces,” {\it Pattern Recognition}, vol. 35, no. 11, pp. 2613–2627, 2002.

[13]	Archambeau, N. Delannay, and M. Verleysen, “Mixtures of robust probabilistic principal component analyzers,” {\it Neurocomputing}, vol. 71, nos. 7–9, pp. 1274–1282, 2008.

[14]	Gruber and Y. Weiss, “Multibody factorization with uncertainty and missing data using the EM algorithm,” in {\it Proceedings of IEEE Conference on Computer Vision and Pattern Recognition}, vol. 1, pp. 707–714, 2004.

[15]	Y. Ma, H. Derksen, W. Hong, and J. Wright, “Segmentation of multivariate mixed data via lossy data coding and compression,” {\it IEEE Transactions on Pattern Analysis and Machine Intelligence}, vol. 29, no. 9, pp. 1546–1562, 2007.

[16]	Y. Yang, S. R. Rao, and Y. Ma, “Robust statistical estimation and segmentation of multiple subspaces,” in {\it Proceedings of IEEE Conference on Computer Vision and Pattern Recognition Workshop}, p. 99, 2006.

[17]	C.-Y. Lu, H. Min, Z.-Q. Zhao, L. Zhu, D.-S. Huang, and S. Yan, “Robust and efficient subspace segmentation via least squares regression,” in {\it Proceedings of European Conference on Computer Vision}, pp. 347–360, 2012.

[18]	R. Vidal and P. Favaro, “Low rank subspace clustering (LRSC),” {\it Pattern Recognition Letter}, vol. 43, pp. 47–61, 2014.

[19]	Lu, J. Tang, M. Lin, L. Lin, S. Yan, and Z. Lin, “Correntropy induced L2 graph for robust subspace clustering,” in {\it Proceedings of IEEE International Conference Computer Vision}, pp. 1801–1808, 2013.

[20]	Y. X. Wang, H. Xu, and C. Leng, “Provable subspace clustering: When LRR meets SSC,” in {\it Proceedings of Neural Information Processing Systems}, pp. 64–72, 2013.

[21]	E. L. Dyer, A. C. Sankaranarayanan, and R. G. Baraniuk, “Greedy feature selection for subspace clustering,” {\it Journal of Machine Learning Research}, vol. 14, no. 1, pp. 2487–2517, 2013.

[22]	Y. Zhang, Z. Sun, R. He, and T. Tan, “Robust subspace clustering via half-quadratic minimization,” in {\it Proceedings of IEEE International Conference Computer Vision}, pp. 3096–3103, 2013.

[23]	Park, C. Caramanis, and S. Sanghavi, “Greedy subspace clustering,” in {\it Proceedings of Neural Information Processing Systems}, pp. 2753–2761, 2014.

[24]	R. Heckel and H. Bölcskei, “Robust subspace clustering via thresholding,” {\it IEEE Transactions on Information Theory}, vol. 61, no. 11, pp. 6320–6342, 2015.

[25]	B. Li, Y. Zhang, Z. Lin, and H. Lu, “Subspace clustering by mixture of Gaussian regression,” in {\it Proceedings of IEEE Conference on Computer Vision and Pattern Recognition}, pp. 2094–2102, 2015.

[26]	You, D. Robinson, and R. Vidal, “Scalable sparse subspace clustering by orthogonal matching pursuit,” in {\it Proceedings of IEEE Conference on Computer Vision and Pattern Recognition}, pp. 3918–3927, 2016.

[27]	C. You, C.-G. Li, D. P. Robinson, and R. Vidal, “Oracle based active set algorithm for scalable elastic net subspace clustering,” in {\it Proceedings of IEEE Conference on Computer Vision and Pattern Recognition}, pp. 3928–3937, 2016.

[28]	Elhamifar and R. Vidal, “Sparse subspace clustering,” in {\it Proceedings of IEEE Conference on Computer Vision and Pattern Recognition}, pp. 2790–2797, 2009.

[29]	Elhamifar and R. Vidal, “Sparse subspace clustering: Algorithm, theory, and applications,” {\it IEEE Transactions on Pattern Analysis and Machine Intelligence}, vol. 35, no. 11, pp. 2765–2781, 2013.

[30]	Liu, Z. Lin, and Y. Yu, “Robust subspace segmentation by low-rank representation,” in {\it Proceedings of International Conference on Machine Learning}, pp. 663–670, 2010.

[31]	Liu, Z. Lin, S. Yan, J. Sun, Y. Yu, and Y. Ma, “Robust recovery of subspace structures by low-rank representation,” {\it IEEE Transactions on Pattern Analysis and Machine Intelligence}, vol. 35, no. 1, pp. 171–184, 2013.

[32]	C. Lu, J. Feng, Z. Lin, T. Mei and S. Yan, "Subspace Clustering by Block Diagonal Representation," {\it IEEE Transactions on Pattern Analysis and Machine Intelligence}, vol. 41, no. 2, pp. 487-501, 2019.

[33]	C. Li, C. You and R. Vidal, "Structured Sparse Subspace Clustering: A Joint Affinity Learning and Subspace Clustering Framework," {\it IEEE Transactions on Image Processing}, vol. 26, no. 6, pp. 2988-3001, 2017.

[34]	J. Shi and J. Malik, "Normalized cuts and image segmentation," {\it IEEE Transactions on Pattern Analysis and Machine Intelligence}, vol. 22, no. 8, pp. 888-905, 2000.

[35]	Y. Lecun, L. Bottou, Y. Bengio and P. Haffner, "Gradient-based learning applied to document recognition," in {\it Proceedings of the IEEE}, vol. 86, no. 11, pp. 2278-2324, 1998.

[36]	J. Hull, “A database for handwritten text recognition research,” {\it IEEE Transactions on Pattern Analysis and Machine Intelligence}, vol. 16, pp. 550–554, 1994.

[37]	S. Georghiades, P. N. Belhumeur, and D. Kriegman, “From few to many: Illumination cone models for face recognition under variable lighting and pose,” {\it IEEE Transactions on Pattern Analysis and Machine Intelligence}, vol. 23, no. 6, pp. 643–660, 2001.

[38]	X. Wang, X. Guo, Z. Lei, C. Zhang and S. Z. Li, "Exclusivity-Consistency Regularized Multi-view Subspace Clustering," in {\it Proceedings of IEEE Conference on Computer Vision and Pattern Recognition}, pp. 1-9, 2017.

[39]	M. Yin, Y. Guo, J. Gao, Z. He and S. Xie, "Kernel Sparse Subspace Clustering on Symmetric Positive Definite Manifolds," in {\it Proceedings of IEEE Conference on Computer Vision and Pattern Recognition}, pp. 5157-5164, 2016.

[40]	Floyd, Robert W, "Algorithm 97: Shortest Path," {\it Communications of the ACM}, vol. 5, no. 6, pp. 345, 1962.

\end{document}